*Research Note*

# The Good Old Davis-Putnam Procedure Helps Counting Models

**Elazar Birnbaum**                                          ELAZAR@CS.HUJI.AC.IL
**Eliezer L. Lozinskii**                                     LOZINSKI@CS.HUJI.AC.IL
*Institute of Computer Science*
*The Hebrew University*
*91904 Jerusalem, Israel*

## Abstract

As was shown recently, many important AI problems require counting the number of models of propositional formulas. The problem of counting models of such formulas is, according to present knowledge, computationally intractable in a worst case. Based on the Davis-Putnam procedure, we present an algorithm, CDP, that computes the exact number of models of a propositional CNF or DNF formula $F$. Let $m$ and $n$ be the number of clauses and variables of $F$, respectively, and let $p$ denote the probability that a literal $l$ of $F$ occurs in a clause $C$ of $F$, then the average running time of CDP is shown to be $O(m^d n)$, where $d = \lceil \frac{-1}{\log_2(1-p)} \rceil$. The practical performance of CDP has been estimated in a series of experiments on a wide variety of CNF formulas.

## 1. Introduction

Given a propositional formula $F$ in CNF or DNF, one may want to know what is the number $\mu(F)$ of its models, that is, assignments of truth values to its variables that satisfy $F$. This problem of *counting models* has numerous applications (Roth, 1996). Many important AI problems require counting models of propositional formulas or are reducible to this problem. Among these problems are: (a) Computing degree of belief in a propositional statement $s$ with respect to a propositional knowledge base $KB$ as a conditional probability of $s$ given $KB$. In the absence of a sufficient statistics this probability can be estimated by $|\mu(KB \cup \{s\})| / |\mu(KB)|$. (b) Inference in Bayesian belief networks. As Roth (1996) showed, the problem of counting models of a propositional formula can be reduced to the problem of computing the probability that a node in a Bayesian belief network is true. (c) Computing the number of solutions of a constraint satisfaction problem. A CNF formula $F$ on $n$ variables $x_1, \ldots, x_n$ may be regarded as a constraint satisfaction problem in which the domain of each variable is $\{0, 1\}$, and each clause $C$ is a relation $R \subseteq \{0, 1\}^n$ of all n-tuples for which at least one literal of $C$ assumes the value 1. Then the problem of computing the number of solutions of such problems is the one of computing $\mu(F)$. (d) Estimating the utility of reasoning with approximate theory $AT$ instead of the original theory $T$. This utility depends on the size of $|\mu(AT)| - |\mu(T)|$ (Roth, 1996).

Another important application is to reasoning with incomplete information (Grove et al., 1994; Lozinskii, 1989, 1995). Indeed, if $F$ describes a real world $W$ faithfully, then given a formula $\phi$ such that neither $\phi$ nor $\neg\phi$ is a logical consequence of $F$, a reasonable assumption is that the more models of $F$ assert $\phi$, the more likely is that $\phi$ is true in $W$.

The problem of counting models is #P-complete in general (Valiant, 1979), and so, according to the present state of the art in the field, computationally intractable in the worst case. Dubois (1991) introduced an algorithm that for any formula $F$ of $n$ variables and $m$ clauses, each clause containing $r$ literals, counts models of $F$ in time $O(m\alpha_r^n)$, $\alpha_r$ being the positive root of the polynomial $y^r - y^{r-1} - \cdots - 1$ ($\alpha_2 \approx 1.62, \alpha_3 \approx 1.84, \alpha_4 \approx 1.93, \ldots$ and $\lim_{r \to \infty} \alpha_r = 2$). Recently, Zhang (1996) presented an algorithm based on similar ideas and with similar time complexity. The time complexity achieved by these algorithms is an important improvement over that of a naive algorithm which checks all assignments of truth values to the variables of a formula $F$ in time $O(m2^n)$.

To satisfy practical applications, one has either to resort to a good *approximation* of $\mu(F)$, or use an algorithm for computing $\mu(F)$ *exactly* with a reasonable *average* time complexity.

Regarding the first alternative, Luby and Veličković (1991) presented a deterministic algorithm for approximating the proportion of truth assignments that satisfy a DNF formula $F$. Let $Pr[F]$ denote this proportion ($Pr[F] = \mu(F)/2^n$ where $n$ is the number of variables of $F$). Given small numbers $\epsilon, \delta > 0$, the algorithm computes an estimate $Y$ of $Pr[F]$ which satisfies

$$(1 - \delta)Pr[F] - \epsilon \leq Y \leq Pr[F] + \epsilon.$$

Let $m$ and $n$ be the number of clauses and variables of $F$, respectively. The running time of the algorithm is estimated by a polynomial in $n$ and $m$ multiplied by

$$n^{O(\frac{1}{\delta}(\log^2 m + \log^2 \frac{1}{\epsilon})(\log \log m + \log \frac{1}{\epsilon} + \log \frac{1}{\delta}))}.$$

So, the running time of the algorithm grows fast with the precision required.

Linial and Nisan (1990) studied the problem of computing the size of a union of a given family of sets $S_1 \cup S_2 \cup \cdots \cup S_m$ by means of the Inclusion-Exclusion formula

$$|\bigcup_{i=1}^{m} S_i| = \sum_{1 \leq i \leq m} |S_i| - \sum_{1 \leq i < j \leq m} |S_i \cap S_j| + \sum_{1 \leq i < j < k \leq m} |S_i \cap S_j \cap S_k| - \cdots$$
$$+ (-1)^{m+1} |S_1 \cap \cdots \cap S_m|. \qquad (1)$$

They showed that the size of the union can be approximated accurately when the sizes of only part of the set intersections are known. Suppose the intersection sizes are known for all subfamilies containing at most $k$ sets. If $k \geq \Omega(\sqrt{m})$, then the union size can be approximated with a relative error of $O(e^{-2k/\sqrt{m}})$.

As Linial and Nisan (1990) pointed out, this result can be applied to approximating the number of models of a DNF formula $F$ — each clause of $F$ has its set of models, and $\mu(F)$ is equal to the size of the union of all those sets. Thus, only the first $k$ terms of the formula (1) are needed for a good approximation, where $\sqrt{m} \leq k \leq m$. However, for a high precision, $k$ approaches $m$. For instance, for $m = 100$ and an error of $10^{-3}$, $k \approx 34$.

Iwama (1989) introduced an algorithm for testing satisfiability of a CNF formula $F$ over $n$ variables by counting the truth assignments falsifying $F$ and checking if their number is less than $2^n$. The counting is accomplished through an Inclusion-Exclusion process, where the size of the set of truth assignments falsifying $F$ is computed with the aid of the observation that this set is the union of the sets of truth assignments falsifying each clause of $F$. Iwama analyzed the average running time of his algorithm on the same framework as we did for the algorithm we present in the sequel. He showed that for a CNF formula satisfying the condition that for a constant $c$, such that $\ln c \geq \ln m - p^2 n$, where $n$ is the number of variables, $m$ is the number of clauses, and $p$ is the probability that a given literal appears in a clause (the same for all literals), the average running time of the algorithm is $O(m^{c+1})$.

Lozinskii (1992) employed an algorithm which is similar to that presented by Iwama (1989), for counting models. He showed that under reasonable assumptions, the average running time of computing the exact number of models of a formula with $m$ clauses on $n$ variables is $O(m^c n)$, where $c = O(\log m)$.

In the sequel we present an algorithm which we call CDP (Counting by Davis-Putnam) for computing $\mu(F)$ precisely. Algorithm CDP is based on the Davis-Putnam procedure (DPP) (Davis & Putnam, 1960) for solving the satisfiability problem (SAT). We study features of DPP, and show how its rules can be modified, resulting in an algorithm for counting models. An analysis shows that for a formula $F$ with $m$ clauses on $n$ variables, the average running time of CDP is $O(m^d n)$, where $d = \lceil \frac{-1}{\log_2(1-p)} \rceil$, and $p$ is the probability that any literal occurs in a given clause.

Section 5 describes numerous experiments with CDP showing that its actual average time complexity is significantly lower than the upper bounds provided by mathematical analysis in Section 3 and in previous works (Dubois, 1991; Iwama, 1989; Lozinskii, 1992; Zhang, 1996). We also point out advantages of this new algorithm over the ones presented in these previous works. Finally, we estimated CDP performance on formulas with 3-literal clauses, and compared the results to those of checking the satisfiability of such formulas by DPP presented by Mitchell et al. (1992).

## 2. Algorithm for Counting Models

In the sequel we consider propositional formulas over a set $V = \{x_1, \ldots, x_n\}$ of variables, $l$ denotes a literal ($l \in \{x_1, \ldots, x_n, \bar{x}_1, \ldots, \bar{x}_n\}$), and $\bar{l}$ stands for the negation of $l$.

### 2.1 The Davis-Putnam Procedure

The Davis-Putnam procedure (DPP) (Davis & Putnam, 1960) is intended for deciding satisfiability of a propositional CNF formula $F$ (regarded as a set of clauses), and can be presented as a recursive boolean function (assuming that no clause of $F$ is a tautology):

function DPP ($F$: propositional CNF formula);

1. if $F$ is empty then
    return *true*;

2. if $F$ contains an empty clause then
    return *false*;

3. (* The pure literal rule *)
    if there exists a pure literal $l$ in $F$ (* such that $\bar{l}$ does not appear in $F$ *) then
        $F_1 = \{C \mid C \in F, l \notin C\}$; (* delete from $F$ all clauses containing $l$ *)
        return DPP($F_1$); (* $F$ is satisfiable iff $F_1$ is so *)

4. (\* The unit clause rule \*)
   if $F$ contains a unit clause $\{l\}$ (\* a clause consisting of a single literal \*) then
   $\quad F_1 = \{C - \{\bar{l}\} \mid C \in F, l \notin C\}$; (\* delete from $F$ $\bar{l}$ and all clauses containing $l$ \*)
   $\quad$ return DPP($F_1$); (\* $F$ is satisfiable iff $F_1$ is so \*)

5. (\* The splitting rule \*)
   choose a variable $x$ of $F$;
   $\quad F_1 = \{C - \{\bar{x}\} \mid C \in F, x \notin C\}$;
   $\quad F_2 = \{C - \{x\} \mid C \in F, \bar{x} \notin C\}$;
   $\quad$ return (DPP($F_1$) $\vee$ DPP($F_2$)). (\* $F$ is satisfiable iff ($F_1 \vee F_2$) is so \*)

It can be seen that the pure literal rule and the unit clause rule are actually particular cases of the splitting rule. If there exists a pure literal $l$ in $F$, and the splitting rule chooses $l$, then $F_1 \subseteq F_2$, and hence ($F_1 \vee F_2$) is satisfiable iff $F_1$ is so. According to the splitting rule, $F$ is satisfiable iff ($F_1 \vee F_2$) is so, and hence, $F$ is satisfiable iff $F_1$ is so. This is exactly the pure literal rule.

If $F$ contains a unit clause $\{l\}$, and the splitting rule chooses $l$, then $F_2$ contains an empty clause (the unit clause without its only literal), and so, is unsatisfiable. So, $F$ is satisfiable iff $F_1$ is so. This is the unit clause rule. (Incidentally, the splitting rule treats well tautological clauses too. If a clause $C$ of $F$ contains a literal $l$ and its complement $\bar{l}$, and the splitting rule is applied to $l$, then $C$ will be erased from both $F_1$ and $F_2$.)

Thus, we may consider DPP as being based on the splitting rule only, while the two other rules serve as guides for choosing a literal efficiently. Indeed, if $F$ contains a unit clause $\{l\}$ or a pure literal $l$, then applying the splitting rule to $l$ yields only *one* formula, but choosing a non-pure and non-unit-clause literal leaves two formulas for further processing.

If $F$ contains neither a unit clause nor a pure literal, then the literal for the splitting rule should be chosen in such a way that produces sets $F_1$ and $F_2$ that are as small as possible. We address this issue in the sequel.

## 2.2 Counting Models

Thinking of counting models of a propositional formula $F$, we notice that by choosing a variable $x$, the splitting rule actually splits the set of models of $F$ into two disjoint subsets: one in which $x$ is true (namely, the models of $F_1$), and the other (models of $F_2$) in which $x$ is false. The chosen variable $x$ appears neither in $F_1$ nor in $F_2$, and there may be other variables that appear in $F$ but not in $F_1$ (or not in $F_2$). These are the variables which belong only to the clauses that contain the literal $x$ (or $\bar{x}$, respectively). Hence, every model of $F_1$ ($F_2$) can be completed to a model of $F$ by assigning *true* (*false*, respectively) to $x$, and *true* or *false* arbitrarily to every variable of $F$ that does not appear in $F_1$ ($F_2$, respectively). Let $\mu(F)$ stand for the number of models of $F$, and $n_1$ ($n_2$) denote the number of variables (other than $x$) which occur in $F$ but not in $F_1$ ($F_2$), then $\mu(F) = 2^{n_1}\mu(F_1) + 2^{n_2}\mu(F_2)$.

As it has been shown in Section 2.1, if checking satisfiability is the goal, and a pure literal (which is not a unit clause) is chosen for splitting, then only $F_1$ should be processed further, while $F_2$ may be disregarded. However, for counting models both $F_1$ and $F_2$ matter. Note that in this case $F_2$ contains the same number of clauses as $F$. So, in contrast to checking satisfiability, the pure literal rule turns out to be inefficient for counting models.

Based on these observations, the following function CDP counts models of a formula $F$ over $n$ variables.

function CDP ($F$: propositional CNF formula; $n$: integer);

1. if $F$ is empty then
       return $2^n$;
2. if $F$ contains an empty clause then
       return $0$;
3. if $F$ contains a unit clause $\{l\}$ then
       $F_1 = \{C - \{\bar{l}\} \mid C \in F, l \notin C\}$;
       return CDP($F_1, n - 1$);
4. choose a variable $x$ of $F$;
       $F_1 = \{C - \{\bar{x}\} \mid C \in F, x \notin C\}$;
       $F_2 = \{C - \{x\} \mid C \in F, \bar{x} \notin C\}$;
       return CDP($F_1, n - 1$) + CDP($F_2, n - 1$).

Any iteration of CDP deletes the chosen variable, and so, reduces the number of variables of $F$ by at least one. This guarantees that CDP terminates (either on step 1 or on step 2).

## 3. Average Running Time

In the analysis of the average running time of the function CDP we follow Goldberg (1979) and Goldberg et al. (1982), who studied the average running time of DPP.

### 3.1 The Recurrence Equation

Let $m$ and $n$ denote the number of clauses and variables of $F$, respectively. Assume that all the literals have the same probability $p$ to appear in each clause $C$ of $F$. (To simplify the analysis we allow $F$ to contain duplicate clauses and tautologies.) Assume also that there is no dependency among the occurrences of different variables and among the occurrences of the same variable in different clauses. For a given distribution of $p$, let $T(m, n)$ denote the average running time of CDP($F, n$) with $F$ containing $m$ clauses.

If $F$ contains $m$ clauses on $n$ variables, then $mn$ bounds the size of $F$. Each step of function CDP can be accomplished during at most two passes over $F$.

**Lemma 1** *If $F$ contains $m$ clauses on $n$ variables, then there is a constant $c$ such that $cmn$ bounds the time of executing one iteration of function CDP.*

It might be thought, that in order to obtain a good bound on the average running time of function CDP, one must show that the unit clause rule is applied frequently, since the splitting rule is the one that introduces the possibility of exponentially long computations. We show that even if only the splitting rule is applied, low bounds are obtained. Therefore, we assume that no use of the unit clause rule is made, and analyze a variant of CDP, in which step 3 is omitted. Obviously, this change does not affect Lemma 1. We also assume that the literal for the splitting rule is chosen randomly.

Let $l$ be the literal that has been chosen for the splitting rule. Since the probability that $l$ occurs in a clause is $p$, the probability that $k$ out of $m$ clauses contain $l$ is $\binom{m}{k}p^k(1-p)^{m-k}$.

The same holds for $\bar{l}$. The number of clauses of formulas $F_1$, $F_2$ produced by splitting $F$ is equal to the number of all clauses of $F$ minus the number of clauses of $F$ which contain the literal $l$, $\bar{l}$, respectively.

**Lemma 2** *Each of formulas $F_1$, $F_2$ produced by splitting $F$ contains $m - k$ clauses with probability $\binom{m}{k}p^k(1-p)^{m-k}$.*

Since each of $F_1$, $F_2$ contains at most $n-1$ variables, the following recurrence inequality for $T(m,n)$ holds.

$$T(m,n) \leq \begin{cases} cmn+ \\ \sum_{k=1}^{m}\binom{m}{k}p^k(1-p)^{m-k}T(m-k,n-1)+ \\ \sum_{k=0}^{m}\binom{m}{k}p^k(1-p)^{m-k}T(m-k,n-1) & n,m \geq 1 \\ \\ 1 & n=0 \text{ or } m=0 \end{cases} \quad (2)$$

The first term ($cmn$) is the time needed for one iteration of CDP. The second and third terms are the expected times needed for computing the number of models of $F_1$ and $F_2$, respectively. Since the literal $l$ appears in $F$, it is impossible that the number of clauses in which $l$ occurs is 0, and so, the summation of the second term starts at 1. The literal $\bar{l}$ may not appear at all in $F$ (if $l$ is pure). Consequently, the summation of the third term begins at 0.

The right-hand side of (2) overestimates $T(m,n)$ since, first, the number of variables of $F_1$, $F_2$ is *at most* $n-1$, and, second, function CDP may stop when $F$ contains an empty clause, even before $m$ or $n$ becomes 0.

To estimate $T(m,n)$, we show, first, that for $p = 1/3$, $T(m,n) = O(m^2n)$, and, second, that for any $p$, $T(m,n) = O(m^dn)$ where $d = \lceil \frac{-1}{\log_2(1-p)} \rceil$.

## 3.2 The Average Running Time for $p = 1/3$

Following Goldberg (1979), we start with the analysis of the average running time for $p = 1/3$. In this case, each variable occurs in each clause positively, negatively or not at all with the same probability (1/3). So, recurrence (2) takes the form

$$T(m,n) \leq \begin{cases} cmn+ \\ (\frac{2}{3})^m T(m,n-1)+ \\ 2\sum_{k=1}^{m}\binom{m}{k}(\frac{1}{3})^k(\frac{2}{3})^{m-k}T(m-k,n-1) & m,n \geq 1 \\ \\ 1 & n=0 \text{ or } m=0 \end{cases} \quad (3)$$

**Theorem 1** *For $p = \frac{1}{3}$, $T(m,n) = O(m^2n)$.*

Proof is given in Appendix A.

## 3.3 The Average Running Time for Any $p$

The assumption of $p = 1/3$ is commonly adopted in probabilistic analysis of algorithms handling CNF or DNF formulas (Franco & Paull, 1983; Goldberg, 1979; Goldberg et al., 1982), which means that for each variable, its occurrence in a clause with or without negation or non-occurrence, all have the same probability. However, in real cases this may not hold, so we analyze the average running time of CDP for any probability $p$ of a literal occurrence in a clause of $F$.

Before analyzing the recurrence equation let us make the following estimation. The function CDP (without the unit clause rule) may be regarded as an algorithm which scans a binary tree. The root of the tree represents the input formula $F$; The children of each internal node represent the formulas obtained as a result of applying the splitting rule to the formula at that internal node, and the leaves of the tree represent empty formulas and formulas which contain an empty clause.

If an internal node represents a formula with $k$ clauses, the expected number of clauses in each of its children is $k(1-p)$. The expected height of the tree $h$ is therefore about $\log_b m$ where $b = \frac{1}{1-p}$. The number of nodes in a complete binary tree of height $h$ is about $2 \cdot 2^h$, and this is the expected number $\eta$ of iterations of the function CDP. So, $\eta = 2 \cdot 2^h = 2m^d$, where $d = \frac{-1}{\log_2(1-p)}$. The running time of each iteration is $O(\bar{m}\bar{n})$ where $\bar{m}$ is the number of clauses and $\bar{n}$ is the number of variables of the formula that is treated at that iteration.

**Theorem 2** $T(m,n) = O(m^d n)$, where $d = \lceil \frac{-1}{\log_2(1-p)} \rceil$.

Proof is given in Appendix A.

A few remarks are in order. First, we assumed that the probability of occurring in a clause is the same for all literals. If this is not the case, then Theorem 2 holds for $p$ being the lowest occurrence probability among all literals of $F$. Second, if $p$ is assumed constant, then the size of clauses, which is of order $2pn$, is proportional to the number of variables, and hence, is supposed to change with $n$.

## 4. Refining the CDP

This section describes some refinements to CDP improving its performance.

### 4.1 Choosing a Variable for the Splitting Rule

The function CDP, like the Davis-Putnam procedure, gives no recommendation how to choose a variable for a split. However, a good choice of this variable may reduce the running time of the function. By 'good choice' we mean one that causes the formulas $F_1$ and $F_2$ to be as small as possible.

The size of these formulas is determined by both $m$ and $n$. As minimizing the number of variables in $F_1$, $F_2$ causes an unjustified computation overhead, we concentrated on reducing the number of clauses $m_1$, $m_2$ in $F_1$, $F_2$, either by minimizing $m_1 + m_2$ (by means of choosing a variable appearing in a maximal number of clauses of $F$), or by minimizing $max(m_1, m_2)$. The latter can be achieved by maximizing over $x$ the quantity $min(pos(x), neg(x))$, where $pos(x)$ $(neg(x))$ denotes the number of clauses of $F$ in which $x$ occurs unnegated (negated, respectively). These two approaches may be combined. Indeed, experiments reported in Section 5 showed best performance if the split variable has been chosen due to $min(m_1 + m_2)$, but if there have been more than one such variable, then a further choice among the competing variables has been made due to $minmax(m_1, m_2)$.

As has been suggested by an anonymous reviewer, by using proper data structures, the computation of $pos(x)$ and $neg(x)$ for all variables $x$ can be done efficiently, causing only a small computation overhead.

## 4.2 Handling Small Formulas

The function CDP stops when $F$ contains an empty clause or when $F$ is empty. What if $F$ consists of only one clause containing $k$ literals? The splitting rule will be applied to $F$, creating $F_1$ and $F_2$ such that one of them may be empty, but the other still consists of one clause involving just one variable less than $F$. So, splitting of the single clause of $F$ will be repeated $k$ times.

A similar process of inefficient splitting takes place also on formulas consisting of relatively few clauses. It turns out that the algorithm presented by Lozinskii (1992) is more efficient for handling small sets of clauses. Experiments on randomly chosen formulas showed that for formulas with less than 6 clauses the algorithm of Lozinskii (1992) performs better than function CDP in its original form. Therefore, when the number of clauses of a formula being processed is reduced under 6, CDP runs the algorithm of Lozinskii (1992).

## 5. Experiments

Algorithm CDP presented in the previous sections has been programmed in TURBO-PASCAL, and run on PC, with the purpose of determining an efficient way of choosing a variable for the splitting rule, finding an appropriate number of clauses in a formula for switching the function CDP to the algorithm of Lozinskii (1992), and assessing the actual running time of CDP.

In the experiments, CDP has been applied to randomly produced sets of clauses with the following parameters: $10 \leq m \leq 200$, $10 \leq n \leq 80$ and $p_1, p_2 \in \{0.1, 0.2, 0.3\}$, where $p_1$, $p_2$ denote the probability of a variable to occur in a clause unnegated or negated, respectively ($p_1$ and $p_2$ are not necessarily the same). For each combination of the parameters, 100 random instances of $F$ have been processed. Performance of CDP has been measured by its running time and by the number of recursive calls to the main function.

We have observed that for each pair of probabilities $p_1, p_2$, there is a threshold number of clauses $\bar{m}$, such that for $m \geq \bar{m}$ the hard cases lie around a corresponding number of variables. For instance, the hard cases for $p_1 = p_2 = 0.1$ and $\bar{m} = 60$ are located around $n = 50$. Figures 1 and 2 present the average number of recursive calls performed by CDP in the hard cases. More data are given in Tables 1-6 of Appendix B. Following a suggestion of one of the reviewers, we checked the standard deviation in all cases (see Table 7 in Appendix B). All the experiments ran on a pentium based PC (266 MHz). The actual running time of CDP ranged from less than one second to 5 hours.

Following the discovery of hard cases for SAT (Mitchell et al., 1992), we ran CDP on a series of random sets of 3-literal clauses with the ratio of $m/n$ from 0.2 to 8.0. For each combination of $m$ and $n$, 200 instances were considered. Figure 3 gives the median number of recursive calls performed by CDP for a sample of $n = 20, 40, 50$. More data in Table 8 of Appendix B. (Median number is shown to enable comparison with the results presented by Mitchell et al., 1992).

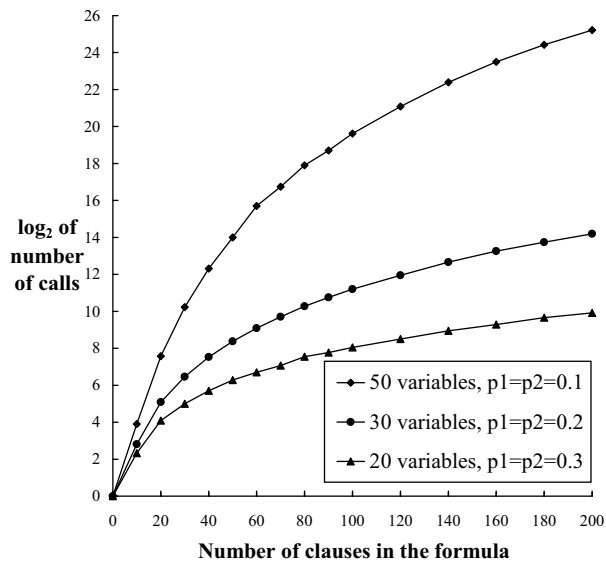

Figure 1: Average number of recursive calls for different values of $p1, p2$

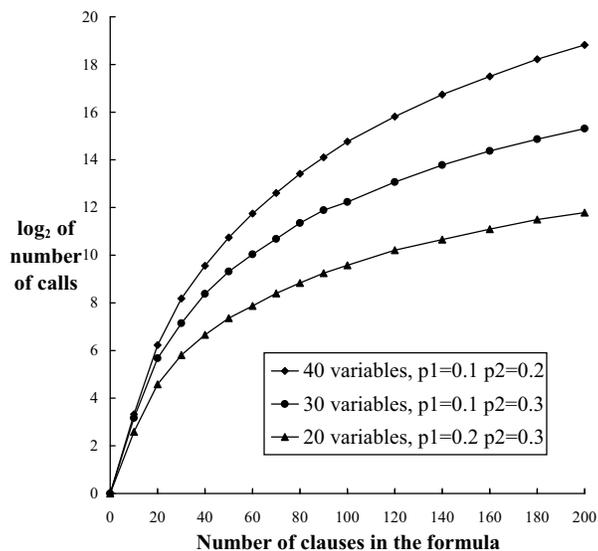

Figure 2: Average number of recursive calls for different values of $p1, p2$ (continued)

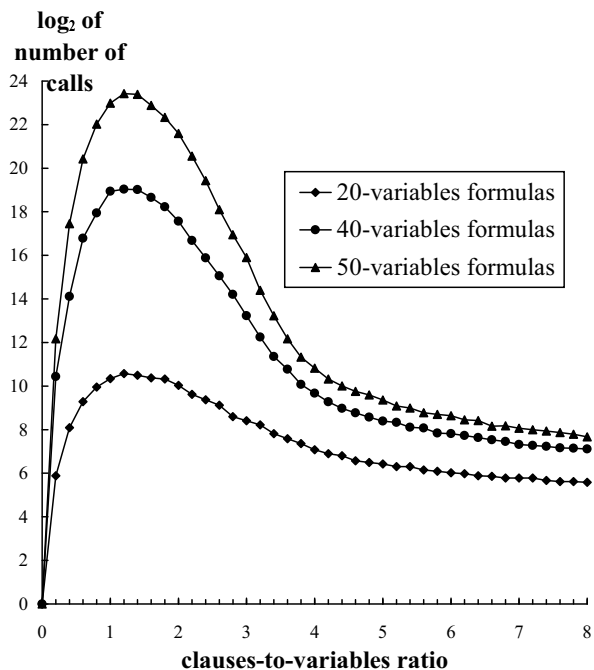

Figure 3: Median number of recursive calls for 3-literal formulas.

In all experiments with counting models we have observed a phenomenon akin to that characteristic of checking satisfiability, namely, for a given length of clauses $r$, the running time of algorithm CDP reaches its maximum at a certain value of the clauses-to-variables ratio. For $r = 3$ (Figure 3) this "hard" ratio is 1.2, while it has been reported in many works (e.g., Mitchell et al., 1992) that hard cases of 3-SAT lie around $m/n = 4.3$. This shift of the maximum running time to the left (from $m/n = 4.3$ to $m/n = 1.2$) can be explained by the fact that when a program for SAT discovers a satisfying assignment, it stops processing, while counting models has to go on over all satisfying assignments. So, the higher is the probability that a formula is satisfiable, and the more models it is likely to have (and this is what happens if the ratio $m/n$ decreases), the longer model counting takes compared to checking satisfiability.

## 6. Concluding Remarks

We have presented an algorithm (CDP) for counting models of a propositional formula $F$. The previous sections considered CNF formulas, however, it is easy to construct a dual algorithm for DNF formulas.

Comparing algorithm CDP with that of Iwama (1989) and Lozinskii (1992), we note that both have a common central feature: the more variables have both negated and unnegated occurrences in the clauses of a given formula, the better is the performance of the algorithms. This feature is also common to the algorithms presented by Dubois (1991) and by Zhang (1996). In contrast to checking satisfiability, pure literals slow down counting models by these algorithms.

The experiments made with algorithm CDP have shown that it performs better than that of Lozinskii (1992). However the main advantage of CDP is in the amount of storage it requires. In the experiments that have been performed by Lozinskii (1992) for $m \leq 100$ and $n \leq 30$, there were cases with thousands of clauses stored during the run of the algorithm. Since algorithm CDP performs a depth-first search of the split tree, and by each split at least one variable is deleted, the maximal number of stored clauses is bounded by $mn$.

During the experiments we checked also the maximal number $M$ of stored clauses. The results showed that $M$ does almost not depend on $n$, such that $M \leq sm^t$, where $t \leq 1.3$ is not depending on $p_1$ and $p_2$, while $s$ depends on $p_1$ and $p_2$. Table 1 displays $s$ as a function of $p_1$ and $p_2$. Figure 1 in Appendix B shows average values of $M$ for $n = 50, p_1 = p_2 = 0.1, 0.2, 0.3$. The maximum of $M$ in all our experiments was 1491 (for $n = 50, m = 200$).

| $p_1$ | $p_2$ | $s$ |
|-------|-------|------|
| 0.1 | 0.1 | 1.54 |
| 0.1 | 0.2 | 0.82 |
| 0.1 | 0.3 | 0.57 |
| 0.2 | 0.2 | 0.74 |
| 0.2 | 0.3 | 0.50 |
| 0.3 | 0.3 | 0.46 |

Table 1: $s$ as a function of $p_1$ and $p_2$

The analysis of the running time of CDP follows the method developed by Goldberg (1979) and by Goldberg et al. (1982). This method was used also by Iwama (1989). The well known criticisms by Franco and Paull (1983), and by Mitchell et al. (1992) points out that the instances generated by this method are mostly satisfiable, and that is indeed the reason for the good performance of satisfiability checking algorithms: the algorithms find rapidly one model of the formula. This point does not apply when we seek not only just one model but all the models.

Finally, it should be noted that the algorithms for counting models developed so far are designed for *propositional* formulas. As it has been mentioned in the introduction, one application of counting models is in reasoning with incomplete information in logic systems usually expressed in *predicate calculus*. Therefore, developing counting algorithms for predicate calculus formulas is a next challenge of a great importance.

## Acknowledgments

We are grateful to the anonymous referees for their most apt and inspiring comments. This research has been partly supported by the Israel Science Foundation administrated by the Israel Academy of Sciences and Humanities.

## Appendix A. Proofs

**Theorem 1** *For $p = \frac{1}{3}$, $T(m, n) = O(m^2 n)$.*

**Proof.** For $p = \frac{1}{3}$, the recurrence inequality takes the form

$$T(m, n) \leq \begin{cases} cmn+ \\ (\frac{2}{3})^m T(m, n-1)+ \\ 2\sum_{k=1}^{m} \binom{m}{k}(\frac{1}{3})^k (\frac{2}{3})^{m-k} T(m-k, n-1) & m, n \geq 1 \\ \\ 1 & n = 0 \text{ or } m = 0 \end{cases} \quad (1)$$

Goldberg (1979) showed, that for the following recurrence, $S(m, n) = O(m^2 n)$.

$$S(m, n) = \begin{cases} cmn+ \\ 2\sum_{k=1}^{m} \binom{m}{k}(\frac{1}{3})^k (\frac{2}{3})^{m-k} S(m-k, n-1) & m, n \geq 1 \\ \\ 1 & n = 0 \text{ or } m = 0 \end{cases} \quad (2)$$

The difference between (2) and (1), the recurrence for $S(m, n)$ and the recurrence for $T(m, n)$, is the extra term $(\frac{2}{3})^m T(m, n-1)$ in the latter. To compare $T(m, n)$ and $S(m, n)$ we first need the following result.

**Lemma 1** *For all $m \geq 0$ and all $n \geq 0$, $S(m, n) \leq (\frac{3}{2})^{m-1} cmn$.*

**Proof.** Goldberg (1979) showed that $S(m, n) \leq cm^2 n$. Since $m < (\frac{3}{2})^{m-1}$ for $m \geq 5$, this proves the lemma for all $m \geq 5$ and all $n \geq 0$. Direct computation of $S(m, n)$, due to (2), for $m = 0, 1, 2, 3, 4$ completes the proof for all $m \geq 0$ and all $n \geq 0$.

Now $T(m, n)$ can be estimated as follows.

**Lemma 2** *For all $m$ and all $n$, $T(m, n) \leq 3 \cdot S(m, n)$.*

**Proof.** By induction on $n$.
The assertion is true for $n \leq 1$.
Suppose it is true for every $i < n$, then

$$\begin{aligned} T(m, n) &\leq cmn + (\frac{2}{3})^m T(m, n-1) + 2\sum_{k=1}^{m} \binom{m}{k}(\frac{1}{3})^k (\frac{2}{3})^{m-k} T(m-k, n-1) \\ &\leq cmn + 3 \cdot (\frac{2}{3})^m S(m, n-1) + \\ &\qquad 3 \cdot 2\sum_{k=1}^{m} \binom{m}{k}(\frac{1}{3})^k (\frac{2}{3})^{m-k} S(m-k, n-1). \end{aligned}$$

According to (2),

$$3S(m, n) = 3cmn + 3 \cdot 2\sum_{k=1}^{m} \binom{m}{k}(\frac{1}{3})^k (\frac{2}{3})^{m-k} S(m-k, n-1),$$

so, $T(m, n) \leq 3S(m, n) + 3(\frac{2}{3})^m S(m, n-1) - 2cmn$. For all $n > 0$, $S(m, n-1) \leq S(m, n)$, and by Lemma 1, for all $m$ and all $n$, $3(\frac{2}{3})^m S(m, n) \leq 2cmn$. This completes the proof.

Since $S(m, n) = O(m^2 n)$ and $T(m, n) \leq 3 \cdot S(m, n)$, we get for $p = \frac{1}{3}$, $T(m, n) = O(m^2 n)$.

**Theorem 2** $T(m, n) = O(m^d n)$, where $d = \lceil \frac{-1}{\log_2(1-p)} \rceil$.

**Proof.** We analyze the recurrence inequality for any $p$, using considerations similar to those of Goldberg et al. (1982).

The recurrence inequality takes the form:

$$T(m, n) \leq \begin{cases} cmn+ \\ (1-p)^m T(m, n-1)+ \\ 2\sum_{k=1}^{m} \binom{m}{k} p^k (1-p)^{m-k} T(m-k, n-1) \qquad m, n \geq 1 \\ \\ 1 \qquad\qquad\qquad\qquad\qquad\qquad\qquad\quad n = 0 \text{ or } m = 0 \end{cases} \qquad (3)$$

It is easy to show, by induction on $n$, that $T(m, n)$ grows with $n$. Hence,

$$T(m, n) \leq cmn + (1-p)^m T(m, n) + 2\sum_{k=1}^{m} \binom{m}{k} p^k (1-p)^{m-k} T(m-k, n). \qquad (4)$$

A solution of (4) is linear in $n$, since this recurrence is linear in the $T's$. Now for any value $m_0$ let us choose a constant $a$ such that for all $m \leq m_0$

$$T(m, n) \leq am^d n, \qquad (5)$$

where $d = \lceil \frac{-1}{\log_2(1-p)} \rceil$. Then for all $m \leq m_0$, (4), (5) imply

$$\begin{aligned} (1 - (1-p)^m) T(m, n) &\leq cmn + 2\sum_{k=1}^{m} \binom{m}{k} p^k (1-p)^{m-k} T(m-k, n) \\ &\leq cmn + 2\sum_{k=1}^{m} \binom{m}{k} p^k (1-p)^{m-k} a(m-k)^d n \\ &= cmn + am^d n \cdot 2\sum_{k=1}^{m} \binom{m}{k} p^k (1-p)^{m-k} (1 - \frac{k}{m})^d. \end{aligned}$$

As Goldberg et al. (1982) showed,

$$\sum_{k} \binom{m}{k} p^k (1-p)^{m-k} (1 - \frac{k}{m})^d = (1-p)^d + O(m^{-1}), \qquad (6)$$

and therefore,

$$(1 - (1-p)^m) T(m, n) \leq cmn + 2am^d n(1-p)^d + an \cdot O(m^{d-1}).$$

$cmn = O(mn)$, and so,

$$T(m, n) \leq \frac{2am^d n(1-p)^d}{1 - (1-p)^m} + an \cdot O(m^{max(1, d-1)}).$$

We are going now to find out conditions under which

$$\frac{2am^d n(1-p)^d}{1 - (1-p)^m} + an \cdot O(m^{max(1, d-1)}) \leq am^d n,$$

that is

$$O(m^{max(1,d-1)}) \leq m^d[1 - \frac{2(1-p)^d}{1-(1-p)^m}]. \tag{7}$$

First, let $g$ denote the subexpression of (7) in square brackets. For all $d$, $g$ is less than 1, hence, $d$ must be greater than 1 to satisfy $O(m^{max(1,d-1)}) \leq m^d$. So,

$$d > 1 \tag{8}$$

Second, for $0 < p < 1$, $(1-p)^m$ diminishes monotonically, and tends to 0 as $m \to \infty$, hence to make $g$ positive, we need

$$d > \frac{-1}{\log_2(1-p)} \tag{9}$$

Thus, choosing $m$ (and hence, $m_0$) sufficiently large (depending on the value of the $O(m^{-1})$ expression in (6)), and determining the values of $a$ and $d$ satisfying conditions (5), (8), (9), establishes $T(m,n) = O(m^d n)$, where $d = \lceil \frac{-1}{\log_2(1-p)} \rceil$.

## Appendix B. More Experimental Data

| Number of clauses | 40 variables | 50 variables | 60 variables |
|:---:|---:|---:|---:|
| 10 | 16 | 15 | 13 |
| 20 | 234 | 191 | 153 |
| 30 | 1600 | 1203 | 868 |
| 40 | 6294 | 5076 | 3699 |
| 50 | 19856 | 16384 | 11370 |
| 60 | 46645 | 53296 | 28467 |
| 70 | 101607 | 109327 | 75472 |
| 80 | 186199 | 244248 | 153384 |
| 90 | 299220 | 426169 | 311668 |
| 100 | 507888 | 807669 | 635712 |
| 120 | 1093383 | 2222499 | 1687147 |
| 140 | 1945874 | 5507340 | 4289713 |
| 160 | 3095848 | 11824325 | 10812931 |
| 180 | 4441760 | 22444086 | 20230352 |
| 200 | 6377501 | 39119045 | 37381207 |

Table 1: Average number of recursive calls for $p_1 = p_2 = 0.1$

| Number of clauses | 20 variables | 30 variables | 40 variables |
|:---:|---:|---:|---:|
| 10 | 8 | 7 | 7 |
| 20 | 41 | 34 | 29 |
| 30 | 110 | 88 | 76 |
| 40 | 234 | 185 | 150 |
| 50 | 433 | 332 | 266 |
| 60 | 648 | 545 | 424 |
| 70 | 973 | 835 | 636 |
| 80 | 1302 | 1240 | 919 |
| 90 | 1683 | 1731 | 1225 |
| 100 | 2089 | 2349 | 1647 |
| 120 | 3128 | 3947 | 2791 |
| 140 | 3934 | 6461 | 4339 |
| 160 | 5209 | 9821 | 6446 |
| 180 | 6051 | 13661 | 9002 |
| 200 | 6513 | 18696 | 12193 |

Table 2: Average number of recursive calls for $p_1 = p_2 = 0.2$

| Number of clauses | 20 variables | 30 variables |
|:---:|---:|---:|
| 10 | 5 | 5 |
| 20 | 17 | 15 |
| 30 | 32 | 29 |
| 40 | 52 | 45 |
| 50 | 78 | 65 |
| 60 | 104 | 89 |
| 70 | 134 | 114 |
| 80 | 176 | 146 |
| 90 | 218 | 178 |
| 100 | 265 | 215 |
| 120 | 363 | 295 |
| 140 | 495 | 384 |
| 160 | 623 | 489 |
| 180 | 807 | 607 |
| 200 | 966 | 724 |

Table 3: Average number of recursive calls for $p_1 = p_2 = 0.3$

| Number of clauses | 30 variables | 40 variables | 50 variables |
|:---:|---:|---:|---:|
| 10 | 11 | 10 | 9 |
| 20 | 89 | 75 | 60 |
| 30 | 343 | 290 | 211 |
| 40 | 1025 | 752 | 572 |
| 50 | 2179 | 1709 | 1252 |
| 60 | 4195 | 3438 | 2473 |
| 70 | 7309 | 6250 | 4311 |
| 80 | 11740 | 10919 | 7622 |
| 90 | 19224 | 17619 | 10999 |
| 100 | 25866 | 27847 | 16734 |
| 120 | 44729 | 57548 | 37051 |
| 140 | 74280 | 109469 | 68354 |
| 160 | 109134 | 185568 | 117175 |
| 180 | 144698 | 304983 | 201822 |
| 200 | 182516 | 463503 | 307851 |

Table 4: Average number of recursive calls for $p_1 = 0.1$, $p_2 = 0.2$

| Number of clauses | 30 variables | 40 variables | 50 variables |
|---|---|---|---|
| 10 | 9 | 9 | 8 |
| 20 | 51 | 41 | 36 |
| 30 | 141 | 114 | 97 |
| 40 | 333 | 239 | 206 |
| 50 | 635 | 440 | 361 |
| 60 | 1048 | 782 | 596 |
| 70 | 1638 | 1117 | 903 |
| 80 | 2603 | 1706 | 1317 |
| 90 | 3801 | 2493 | 1915 |
| 100 | 4813 | 3523 | 2568 |
| 120 | 8591 | 6087 | 4401 |
| 140 | 14043 | 10221 | 7254 |
| 160 | 21237 | 15087 | 10731 |
| 180 | 29782 | 22832 | 15271 |
| 200 | 40550 | 31882 | 20760 |

Table 5: Average number of recursive calls for $p_1 = 0.1$, $p_2 = 0.3$

| Number of clauses | 20 variables | 30 variables |
|---|---|---|
| 10 | 6 | 6 |
| 20 | 24 | 21 |
| 30 | 56 | 45 |
| 40 | 101 | 80 |
| 50 | 164 | 124 |
| 60 | 234 | 189 |
| 70 | 337 | 255 |
| 80 | 456 | 352 |
| 90 | 606 | 447 |
| 100 | 762 | 567 |
| 120 | 1183 | 851 |
| 140 | 1608 | 1225 |
| 160 | 2184 | 1621 |
| 180 | 2887 | 2159 |
| 200 | 3533 | 2743 |

Table 6: Average number of recursive calls for $p_1 = 0.2$, $p_2 = 0.3$

| Probabilities | Variables | Clauses | Average | Standard deviation |
|---|---|---|---|---|
| $p_1 = p_2 = 0.1$ | 40 | 120 | 1093383 | 412246 |
| | | 200 | 6377501 | 2921029 |
| | 50 | 120 | 2222499 | 1141909 |
| | | 200 | 39119045 | 14587096 |
| $p_1 = p_2 = 0.2$ | 30 | 120 | 3947 | 906 |
| | | 200 | 18696 | 3387 |
| | 40 | 120 | 2791 | 473 |
| | | 200 | 12193 | 2097 |
| $p_1 = p_2 = 0.3$ | 20 | 120 | 363 | 43 |
| | | 200 | 966 | 106 |
| | 30 | 120 | 295 | 25 |
| | | 200 | 724 | 63 |
| $p_1 = 0.1,\ p_2 = 0.2$ | 40 | 120 | 57548 | 21946 |
| | | 200 | 463503 | 135171 |
| | 50 | 120 | 37051 | 11081 |
| | | 200 | 307851 | 90897 |
| $p_1 = 0.1,\ p_2 = 0.3$ | 30 | 120 | 8591 | 2244 |
| | | 200 | 40550 | 7003 |
| | 40 | 120 | 6087 | 1349 |
| | | 200 | 31882 | 6264 |
| $p_1 = 0.2,\ p_2 = 0.3$ | 20 | 120 | 1183 | 183 |
| | | 200 | 3533 | 514 |
| | 30 | 120 | 851 | 126 |
| | | 200 | 2743 | 392 |

Table 7: Samples of standard deviation of number of recursive calls

| Clauses-to-variables ratio | 20 variables | 40 variables | 50 variables |
|---|---|---|---|
| 0.2 | 59 | 1056 | 32062 |
| 0.4 | 272 | 17715 | 2500987 |
| 0.6 | 625 | 113185 | 11634670 |
| 0.8 | 993 | 252044 | 31376686 |
| **1.0** | 1301 | 502669 | 34843025 |
| 1.2 | 1521 | 538579 | 59443580 |
| 1.4 | 1445 | 529520 | 48805440 |
| 1.6 | 1328 | 412563 | 26416134 |
| 1.8 | 1283 | 307939 | 29494057 |
| **2.0** | 1050 | 194754 | 11792351 |
| 2.2 | 789 | 105089 | 8277529 |
| 2.4 | 664 | 60548 | 4037986 |
| 2.6 | 558 | 34057 | 2208147 |
| 2.8 | 387 | 18875 | 923228 |
| **3.0** | 341 | 9574 | 633407 |
| **4.0** | 135 | 816 | 21339 |
| **5.0** | 86 | 335 | 1443 |
| **6.0** | 65 | 225 | 862 |
| **7.0** | 55 | 160 | 558 |
| **8.0** | 48 | 138 | 476 |

Table 8: Median number of recursive calls for 3-literal clauses, $n = 20, 40, 50$

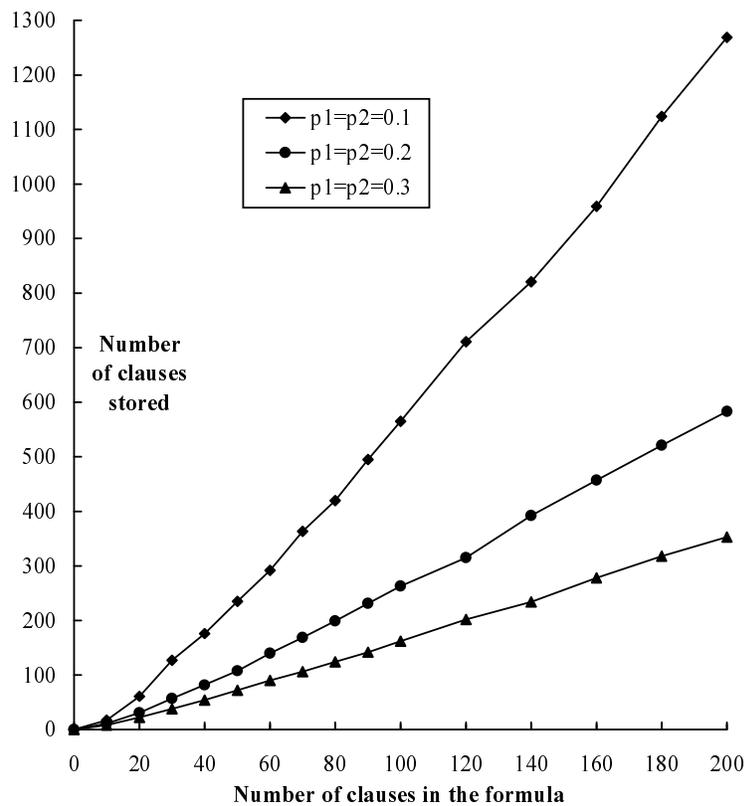

Figure 1: Average of maximal number of clauses stored during the running of CDP for 50 variables.

## References

Davis, M., & Putnam, H. (1960). A computing procedure for quantification theory. *Journal of the ACM*, *7*, 201–215.

Dubois, J. (1991). Counting the number of solutions for instances of satisfiability. *Theoretical Computer Science*, *81*(1), 49–64.

Franco, J., & Paull, M. (1983). Probabilistic analysis of the Davis-Putnam procedure for solving the satisfiability problem. *Discrete Applied Mathematics*, *5*, 77–87.

Goldberg, A. (1979). Average case complexity of the satisfiability problem. In *Proceedings of the 4th Workshop on Automated Deduction* Austin, Texas.

Goldberg, A., Purdom, P., & Brown, C. (1982). Average time analysis of simplified Davis-Putnam procedures. *Information Processing Letters*, *15*(2), 72–75.

Grove, A., Halpern, J., & Koller, D. (1994). Random worlds and maximum entropy. *Journal of Artificial Intelligence Research*, *2*, 33–88.

Iwama, K. (1989). CNF satisfiability test by counting and polynomial average time. *SIAM Journal on Computing*, *18*(2), 385–391.

Linial, N., & Nisan, N. (1990). Approximate inclusion-exclusion. *Combinatorica*, *10*(4), 349–365.

Lozinskii, E. (1989). Answering atomic queries in indefinite deductive databases. *International Journal of Intelligent Systems*, *4*(4), 403–429.

Lozinskii, E. (1992). Counting propositional models. *Information Processing Letters*, *41*(6), 327–332.

Lozinskii, E. (1995). Is there an alternative to parsimonious semantics ?. *Journal of Experimental and Theoretical Artificial Intelligence*, *7*, 361–378.

Luby, M., & Veličković, B. (1991). On deterministic approximation of DNF. In *Proceedings of STOC'91*.

Mitchell, D., Selman, B., & Levesque, H. (1992). Hard and easy distributions of SAT problems. In *Proceedings of AAAI-92*.

Roth, D. (1996). On the hardness of approximate reasoning. *Artificial Intelligence*, *82*, 273–302.

Valiant, L. (1979). The complexity of computing the permanent. *Theoretical Computer Science*, *8*, 189–201.

Zhang, W. (1996). Number of models and satisfiability of sets of clauses. *Theoretical Computer Science*, *155*(1), 277–288.